\documentclass[conference, 10pt]{IEEEtran}

\ifCLASSINFOpdf
\else
\fi

\usepackage{graphicx}
\usepackage{caption}
\usepackage{subfigure}
\usepackage{color}
\usepackage{placeins}
\usepackage{float}
\usepackage{tabularx,colortbl}
\usepackage{times,amsmath,epsfig}
\usepackage{epstopdf}
\usepackage{mathrsfs}
\usepackage{diagbox}
\usepackage{multirow}
\usepackage{algorithm}
\usepackage{algorithmic}

\begin{document}

\title{Deep Convolutional AutoEncoder-based Lossy Image Compression}

\author{\IEEEauthorblockN{Zhengxue Cheng\IEEEauthorrefmark{1}, Heming Sun, Masaru Takeuchi\IEEEauthorrefmark{1}, and Jiro Katto\IEEEauthorrefmark{1}}%
\IEEEauthorblockA{\IEEEauthorrefmark{1}Graduate School of Fundamental Science and Engineering, Waseda University, Tokyo, Japan}
Email: zxcheng@asagi.waseda.jp, terrysun1989@akane.waseda.jp, masaru-t@aoni.waseda.jp, katto@waseda.jp
}

\maketitle

\begin{abstract}
Image compression has been investigated as a fundamental research topic for many decades. Recently, deep learning has achieved great success in many computer vision tasks, and is gradually being used in image compression. In this paper, we present a lossy image compression architecture, which utilizes the advantages of convolutional autoencoder (CAE) to achieve a high coding efficiency. First, we design a novel CAE architecture to replace the conventional transforms and train this CAE using a rate-distortion loss function. Second, to generate a more energy-compact representation, we utilize the principal components analysis (PCA) to rotate the feature maps produced by the CAE, and then apply the quantization and entropy coder to generate the codes. Experimental results demonstrate that our method outperforms traditional image coding algorithms, by achieving a 13.7\% BD-rate decrement on the Kodak database images compared to JPEG2000. Besides, our method maintains a moderate complexity similar to JPEG2000.
\end{abstract}
\IEEEpeerreviewmaketitle

\linespread{0.9}
\vspace{-2mm}
\section{Introduction}
\vspace{-2mm}

Image compression has been a fundamental and significant research topic in the field of image processing for several decades. Traditional image compression algorithms, such as JPEG~\cite{IEEEexample:JPEG} and JPEG2000~\cite{IEEEexample:JPEG2000}, rely on the hand-crafted encoder/decoder (codec) block diagram. They use the fixed transform matrixes, i.e. Discrete cosine transform (DCT) and wavelet transform, together with quantization and entropy coder to compress the image. However, they are not expected to be an optimal and flexible image coding solution for all types of image content and image formats.

Deep learning has been successfully applied in various computer vision tasks and has the potential to enhance the performance of image compression. Especially, the autoencoder has been applied in dimensionality reduction, compact representations of images, and generative models learning~\cite{IEEEexample:autoencoder}. Thus, autoencoders are able to extract more compressed codes from images with a minimized loss function, and are expected to achieve better compression performance than existing image compression standards including JPEG and JPEG2000. Another advantage of deep learning is that although the development and standardization of a conventional codec has historically taken years, a deep learning based image compression approach can be much quicker with new media contents and new media formats, such as 360-degree image and virtual reality (VR)~\cite{IEEEexample:Theis}. Therefore, deep learning based image compression is expected to be more general and more efficient.

Recently, some approaches have been proposed to take advantage of the autoencoder for image compression. Due to the inherent non-differentiability of round-based quantization, a quantizer cannot be directly incorporated into autoencoder optimization. Thus, the works~\cite{IEEEexample:Theis} and~\cite{IEEEexample:Balle} proposed a differentiable approximation for quantization and entropy rate estimation for an end-to-end training with gradient backpropagation. Unlike those works, the work~\cite{IEEEexample:Toderici01} used an LSTM recurrent network for compressing small thumbnail images ($32\times32$), and used a binarization layer to replace the quantization and entropy coder. This approach was further extended in~\cite{IEEEexample:Toderici} for compressing full-resolution images. These works achieved promising coding performance; however, there is still room for improvement, because they did not analyze the energy compaction property of the generated feature maps and did not use a real entropy coder to generate the final codes.

In this paper, we propose a convolutional autoencoder (CAE) based lossy image compression architecture. Our main contributions are twofold.
\begin{itemize}
\item[1)] To replace the transform and inverse transform in traditional codecs, we design a symmetric CAE structure with multiple downsampling and upsampling units to generate feature maps with low dimensions. We optimize this CAE using an approximated rate-distortion loss function.
\item[2)] To generate a more energy-compact representation, we propose a principal components analysis (PCA)-based rotation to generate more zeros in the feature maps. Then, the quantization and entropy coder are utilized to compress the data further.
\end{itemize}
Experimental results demonstrate that our method outperforms JPEG and JPEG2000 in terms of PSNR, and achieves a 13.7\% BD-rate decrement compared to JPEG2000 with the popular Kodak database images. In addition, our method is computationally more appealing compared to other autoencoder based image compression methods.

The rest of this paper is organized as follows. Section II presents the proposed CAE based image compression architecture, which includes the design of the CAE network architecture, quantization, and entropy coder. Section III summarizes the experimental results and compares the rate-distortion (RD) curves of the proposed CAE with those of existing codecs. Conclusion and future work are given in Section IV.

\begin{figure*}[htb]
	\centerline{\psfig{figure=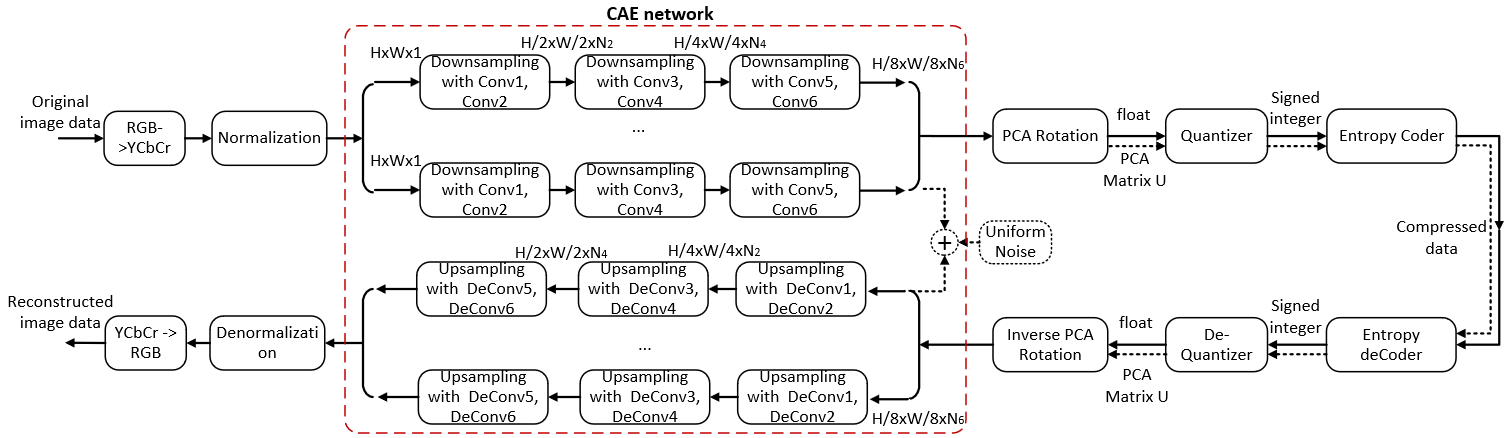,width=178.7mm} }
	\caption{Block diagram of the proposed CAE based image compression. (The detailed block for downsampling/upsampling is shown in Fig. \ref{fig:CAEpair})}
	\label{fig:overall}
\end{figure*}

\section{Proposed Convolutional Autoencoder based Image Compression}

The block diagram of the proposed image compression based on CAE is illustrated in Fig.\ref{fig:overall}. The encoder part includes the pre-processing steps, CAE computation, PCA rotation, quantization, and entropy coder. The decoder part mirrors the architecture of the encoder.

To build an effective codec for image compression, we train this approach in two stages. First, a symmetric CAE network is designed using convolution and deconvolution filters. Then, we train this CAE greedily using an RD loss function with an added uniform noise, which is used to imitate the quantization noises during the optimizing process. Second, by analyzing the produced feature maps from the pre-trained CAE, we utilize the PCA rotation to produce more zeros for improving the coding efficiency further. Subsequently, quantization and entropy coder are used to compress the rotated feature maps and the side information for PCA (matrix U) to generate the compressed bitstream. Each of these components will be discussed in detail in the following.

\vspace{-2mm}
\subsection{CAE Network}

As the pre-processing steps before the CAE design, the raw RGB image is mapped to YCbCr images and normalized to [0,1]. For general purposes, we design the CAE for each luma or chroma component; therefore, the CAE network handles inputs of size $H\times W\times 1$. When the size of raw image is larger than $H\times W$, the image will be split into non-overlapping $H\times W$ patches, which can be compressed independently.

The CAE network can be regarded as an analysis transform with the encoder function, $y=f_{\theta}(x)$, and a synthesis transform with the decoder function, $\hat{x}=g_{\phi}(y)$, where $x$, $\hat{x}$, and $y$ are the original images, reconstructed images, and the compressed data, respectively. $\theta$ and $\phi$ are the optimized parameters in the encoder and decoder, respectively.

\begin{figure}[tb]
	\centerline{\psfig{figure=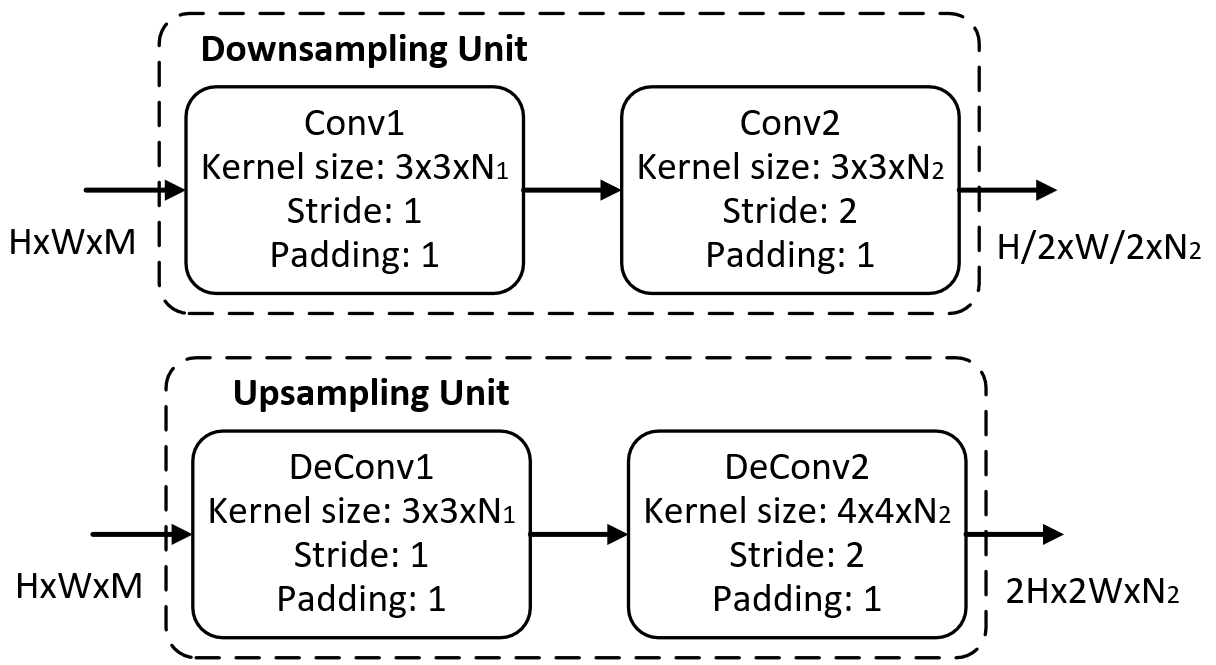,width=60.7mm} }
	\caption{Downsampling/Upsampling Units with two (De)Convolution Filters.}
	\label{fig:CAEpair}
\end{figure}

To obtain the compressed representation of the input images, downsampling/upsampling operations are required in the encoding/decoding process of CAE. However, consecutive downsampling operations will reduce the quality of the reconstructed images. In the work~\cite{IEEEexample:Theis}, it points out that the super resolution is achieved more efficiently by first convolving images and then upsampling them. Therefore, we propose a pair of convolution/deconvolution filters for upsampling or downsampling, as shown in Fig. \ref{fig:CAEpair}, where $N_{i}$ denotes the number of filters in the convolution or deconvolution block. By setting the stride as 2, we can get downsampled feature maps. The padding size is set as one to maintain the same size as the input. Unlike the work~\cite{IEEEexample:Theis}, we do not use residual networks and sub-pixel convolutions, instead, we apply deconvolution filters to achieve a symmetric and simple CAE network.

In traditional codecs, the quantization is usually implemented using the round function (denoted as $[\cdot]$), and the derivative of the round function is almost zero except at the integers. Due to the non-differentiable property of rounding function, the quantizer cannot be directly incorporated into the gradient-based optimization process of CAE. Thus, some smooth approximations are proposed in related works. Theis et al.~\cite{IEEEexample:Theis} proposed to replace the derivative in the backward pass of back propagation as $\frac{d}{dy}([y]) \approx 1$. Balle et al.~\cite{IEEEexample:Balle} replaced the quantization by an additive uniform noise as $[y] \approx y+\mu$. Toderici et al.~\cite{IEEEexample:Toderici01} used a stochastic binarization function as $b(y)=-1$ when $y<0$, and $b(y)=1$ otherwise. In our method, we use the simple uniform noises intuitively to imitate the quantization noises during the CAE training. After CAE training, we apply the real round-based quantization in the final image compression. The network architecture of CAE is shown in Fig. \ref{fig:overall}, in which $N_{i}$ denotes the number of filters in each convolution layer and determines the number of generated feature maps.

\begin{figure}[tb]
	\centerline{\psfig{figure=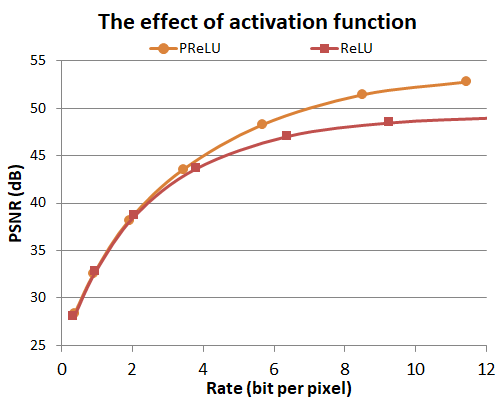,width=50.7mm} }
	\caption{The effect of activation function in CAE.}
	\label{fig:relu}
\end{figure}

As for the activation function in each convolution layer, we utilize the Parametric Rectified Linear Unit (PReLU) function~\cite{IEEEexample:kaiming}, instead of the ReLU which is commonly used in the related works. The performance with ReLU and PReLU functions are shown in Fig.~\ref{fig:relu}. Compared to ReLU, PReLU can improve the quality of the reconstructed images, especially for high bit rate. Inspired by the rate-distortion cost function in the traditional codecs, the loss function of CAE is defined as
\begin{equation}
\begin{aligned}
J(\theta, \phi; x) &= ||x-\hat{x}||^{2}+\lambda\cdot||y||^{2}\\
&= ||x-g_{\phi}(f_{\theta}(x)+\mu)||^{2}+\lambda\cdot||f_{\theta}(x)||^{2}\\
\end{aligned}
\end{equation}
where $||x-\hat{x}||^{2}$ denotes the mean square error (MSE) distortion between the original images $x$ and reconstructed images $\hat{x}$. $\mu$ is the uniform noise. $\lambda$ controls the tradeoff between the rate and distortion. $||f_{\theta}(x)||^{2}$ denotes the amplitude of the compressed data $y$, which reflects the number of bits used to encode the compressed data. In this work, the CAE model was optimized using Adam~\cite{IEEEexample:adam}, and was applied to images with the size of $H\times W$. We used a batch size of 16 and trained the model up to $8\times10^{5}$ iterations, but the model reached convergence much earlier. The learning rate was kept at a fixed value of $0.0001$, and the momentum was set as $0.9$ during the training process.

\subsection{PCA Rotation, Quantization, and Entropy Coder}

\begin{figure}[htb]
\centering
\subfigure[]{
\label{Fig.sub.1}
\includegraphics[width=0.09\textwidth]{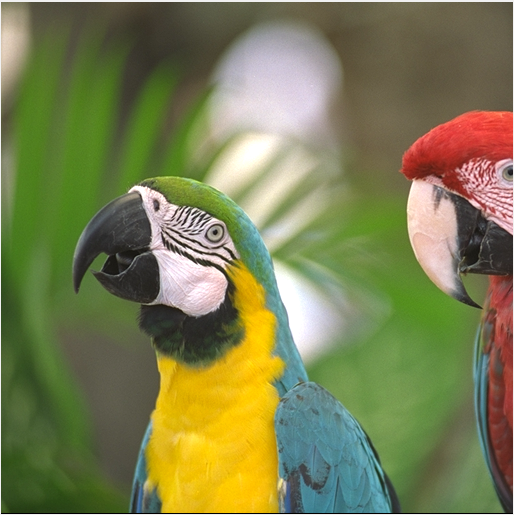}}
\subfigure[]{
\label{Fig.sub.2}
\includegraphics[width=0.18\textwidth]{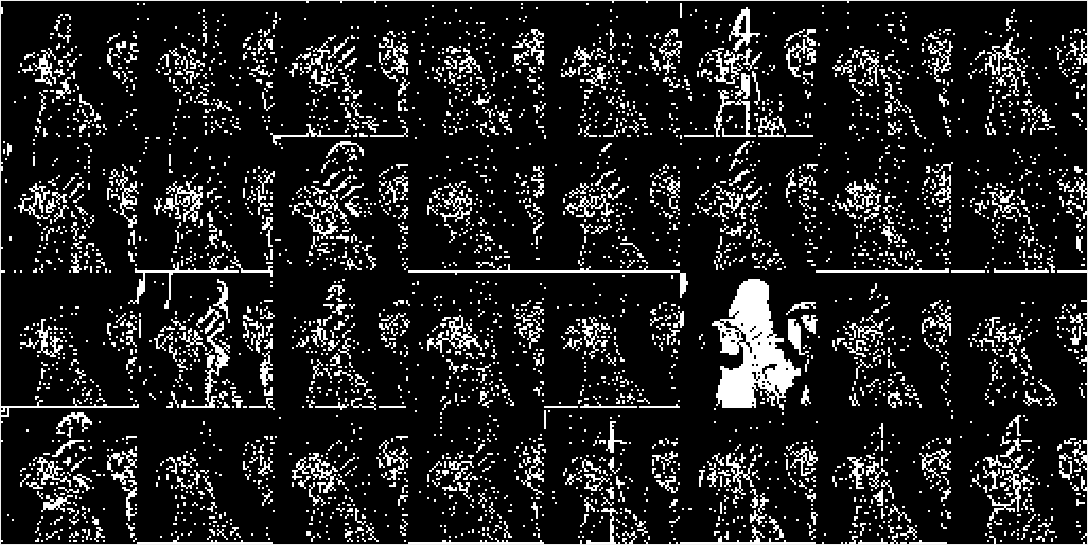}}
\subfigure[]{
\label{Fig.sub.9}
\includegraphics[width=0.18\textwidth]{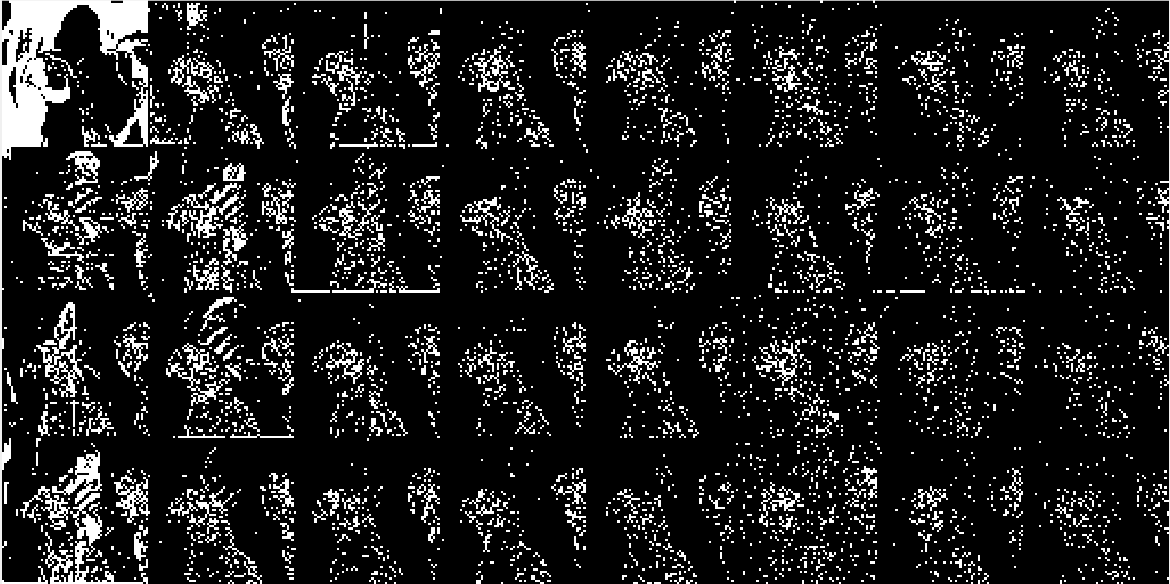}}
\subfigure[]{
\label{Fig.sub.3}
\includegraphics[width=0.09\textwidth]{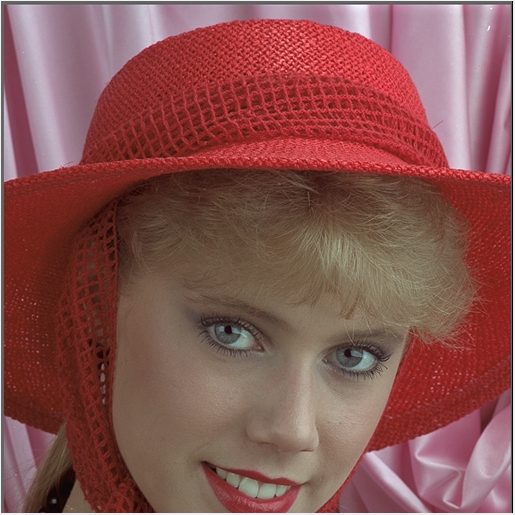}}
\subfigure[]{
\label{Fig.sub.4}
\includegraphics[width=0.18\textwidth]{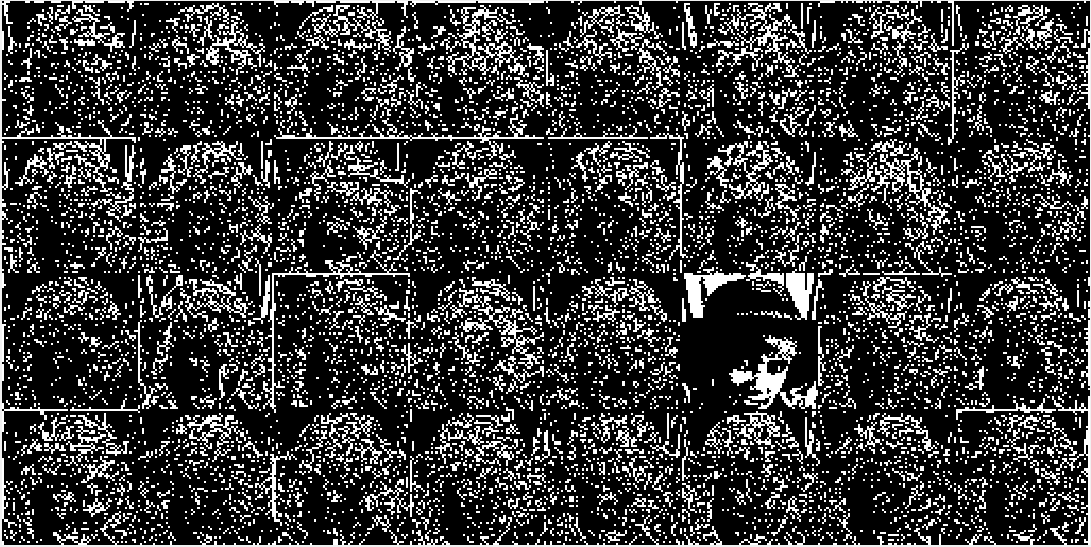}}
\subfigure[]{
\label{Fig.sub.10}
\includegraphics[width=0.18\textwidth]{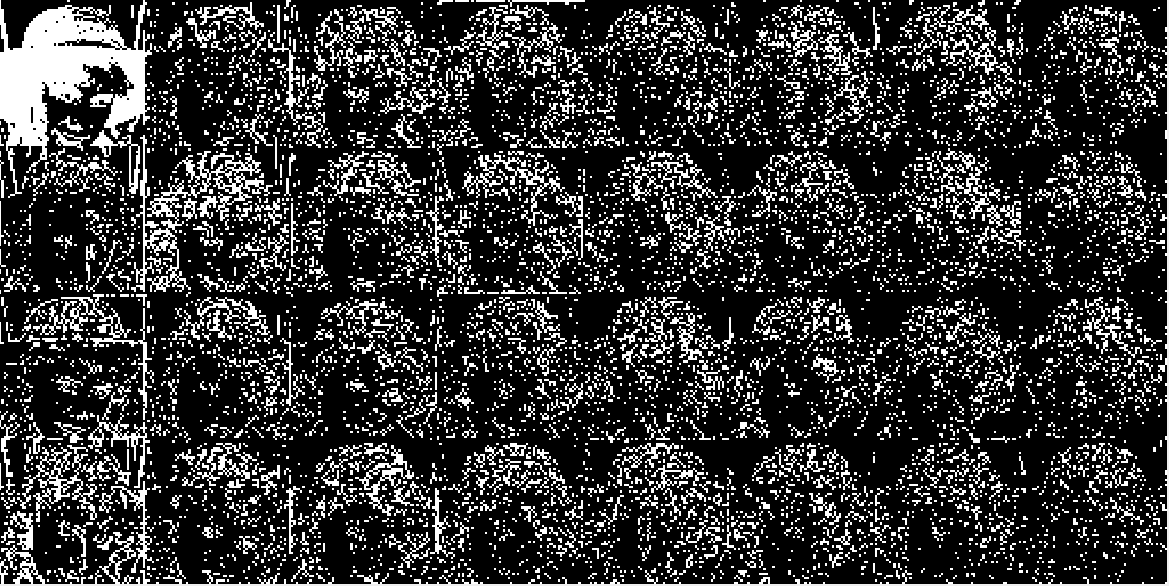}}
\subfigure[]{
\label{Fig.sub.7}
\includegraphics[width=0.09\textwidth]{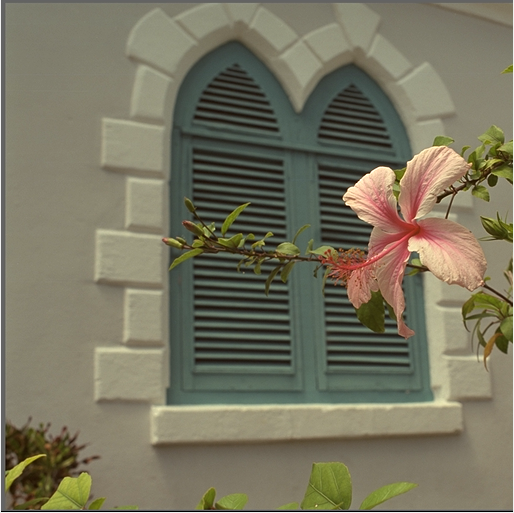}}
\subfigure[]{
\label{Fig.sub.8}
\includegraphics[width=0.18\textwidth]{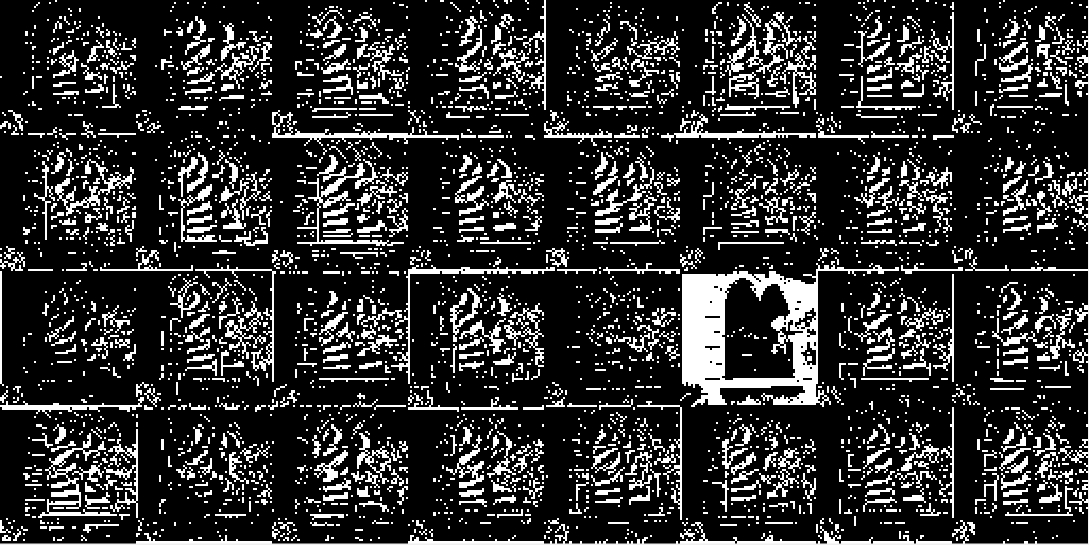}}
\subfigure[]{
\label{Fig.sub.10}
\includegraphics[width=0.18\textwidth]{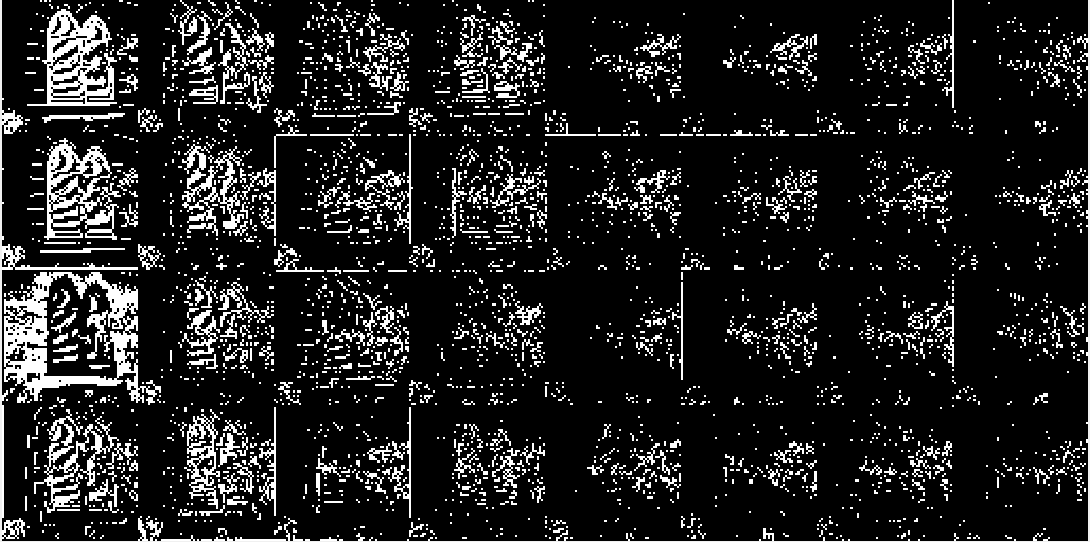}}
\caption{Examples of three images and their corresponding feature maps arranged in raster-scan order ($N_{6}=32$): (a)(d)(g) Raw images, (b)(e)(h) Generated 32 feature maps for Y-component by CAE, and the size of each feature map is $\frac{H}{8}\times\frac{W}{8}$, (c)(f)(i) Rotated Y feature maps by PCA, arranged in vertical scan order.}
\label{fig:features}
\end{figure}

After the CAE computation, an image representation with a size of $\frac{H}{8}\times\frac{W}{8}\times N_{6}$ is obtained for each $H\times W\times 1$ input, where $N_{6}$ denotes the number of filters in the sixth convolution layer of the encoder part. Three examples of the feature maps for the $512\times512$ images cropped from Kodak databases~\cite{IEEEexample:kodak} are demonstrated in the second column of Fig.~\ref{fig:features}. It can be observed that each feature map can be regarded as one high-level representation of the raw images.

To obtain a more energy-compact representation, we decorrelate each feature map by utilizing the principle component analysis (PCA), because PCA is an unsupervised dimensionality reduction algorithm and is suitable for learning the reduced features as a supplementary of CAE. The generated feature maps are denoted as $y = \frac{H}{8}\times\frac{W}{8}\times N_{6}$, and $y$ is reshaped as $N_{6}$-dimensional data. PCA is performed using the following steps. The first step is to compute the covariance matrix of $z$ as follows:
\begin{equation}
\Sigma = \frac{1}{m}\sum_{1}^{m}(y)(y)^{T}
\end{equation}
where $m$ is the number of samples for $y$. The second step is to compute the eigenvectors of $\Sigma$ and stack the eigenvectors in columns to form the matrix $U$. Here, the first column is the principal eigenvector corresponding to the largest eigenvalue, the second column is the second eigenvector, and so on. The third step is to rotate the $N_{6}$-dimensional data $y$ by computing
\begin{equation}
y_{rot} = U^{T}y
\end{equation}

By computing $y_{rot}$, we can ensure that the first feature maps have the largest value, and the features maps are sorted in descending order. Experimental results demonstrate that the vertical-scan order for the feature maps works a little better than diagonal scan and horizontal scan; therefore, we arrange the feature maps in vertical scan as shown in the third column of Fig.~\ref{fig:features}. It can be observed that more zeros are generated in the bottom-right corner and large values are centered in the top-left corner in the rotated feature maps, which can benefit the entropy coder to achieve large compression ratio.

After the PCA rotation, the quantization is performed as
\begin{equation}
y' = [2^{B-1}\cdot y_{rot}]
\end{equation}
where $B$ denotes the number of bits for the desired precision, which is set as 12 in our model.

As for the entropy coder, we use the JPEG2000 entropy coder to decompose $y'$ into bitplanes and apply the adaptive binary arithmetic coder. It is noted that JPEG2000 entropy coder applies EBCOT (Embedded block coding with optimized truncation) algorithm to achieve a desired rate $R$, which is also referred to as post-compression RD optimization. In our method, the feature maps rotated by PCA have many zeros; therefore, assigning the target bits $R$ can further improve the coding efficiency.

In the decoder part, de-quantization is performed as
\begin{equation}
\tilde{y} = \frac{y'}{2^{B-1}}
\end{equation}

After obtaining the float-point number $\tilde{y}$ from the bitstream, we recover the feature maps from the rotated data by using
\begin{equation}
\hat{y} = U\tilde{y}
\end{equation}

Then, the CAE decoder network will reconstruct the images using $\hat{x} = g_{\phi}(\hat{y})$. The side information of PCA rotation is the matrix $U$ with a dimension of $N_{6}\times N_{6}$ for each image. We also quantize $U$ and encode it. The bits for $U$ is added to the final rate as the side information in the experimental results.

\section{Experimental Results}

\subsection{Experimental Setup}

We use a subset of the ImageNet database~\cite{IEEEexample:ImageNet} consisting of 5500 images to train the CAE network. In our experiments, H and W are set as 128; therefore, the images that are input to the CAE are split to a size of $128\times 128$ patches. The numbers of filters, i.e. $N_{i}, i\in[1,6]$ in convolutional layers are set as $\{32,32,64,64,64,32\}$, respectively. The decoder part mirrors the encoder part. The luma component is used to train the CAE network. Mean square error is used in the loss function during the training process in order to measure the distortion between the reconstructed images and original images. For testing, we use the commonly used Kodak lossless image database~\cite{IEEEexample:kodak} with 24 uncompressed $768\times 512$ or $512\times 768$ images. In our CAE training process, $\lambda$ is set as one and the uniform noise $\mu$ is set as $[-\frac{1}{2^{10}}, \frac{1}{2^{10}}]$.

In order to measure the coding efficiency of the proposed CAE-based image compression method, the rate is measured in terms of bit per pixel (bpp). The quality of the reconstructed images is measured using the quality metrics PSNR and MS-SSIM~\cite{IEEEexample:msssim}, which measure the objective quality and perceived quality, respectively.

\subsection{Coding Efficiency Performance}

\begin{figure}[htb]
\centering
\subfigure{
\label{Fig.sub.1}
\includegraphics[width=0.23\textwidth]{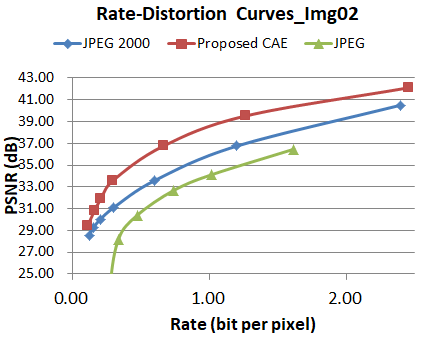}}
\subfigure{
\label{Fig.sub.2}
\includegraphics[width=0.23\textwidth]{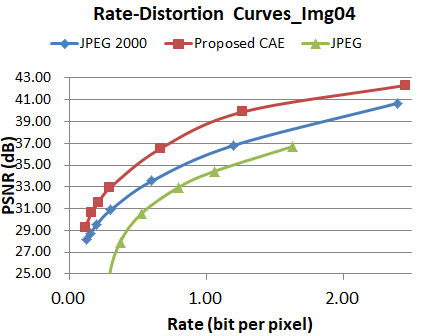}}
\subfigure{
\label{Fig.sub.3}
\includegraphics[width=0.23\textwidth]{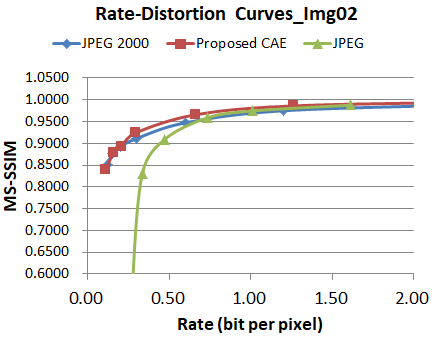}}
\subfigure{
\label{Fig.sub.4}
\includegraphics[width=0.23\textwidth]{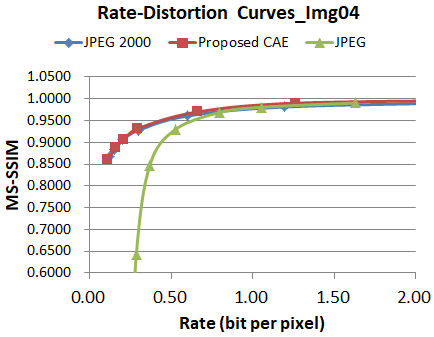}}
\caption{RD curves of color images for the proposed CAE, JPEG, and JPEG2000}
\label{fig:RD}
\end{figure}

\begin{figure}[htb]
\centering
\subfigure{
\label{Fig.sub.1}
\includegraphics[width=0.11\textwidth]{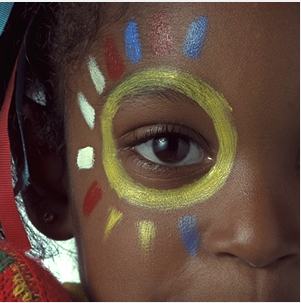}}
\subfigure{
\label{Fig.sub.2}
\includegraphics[width=0.11\textwidth]{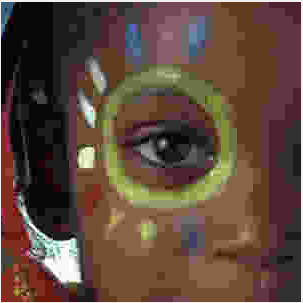}}
\subfigure{
\label{Fig.sub.3}
\includegraphics[width=0.11\textwidth]{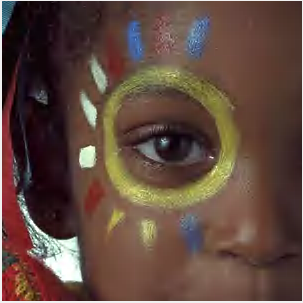}}
\subfigure{
\label{Fig.sub.4}
\includegraphics[width=0.11\textwidth]{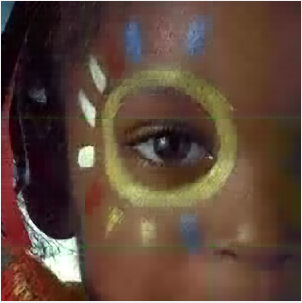}}
\footnotesize{24bpp \qquad\qquad\quad 0.290bpp \qquad\qquad 0.297bpp \qquad \qquad 0.293bpp}
\subfigure{
\label{Fig.sub.1}
\includegraphics[width=0.11\textwidth]{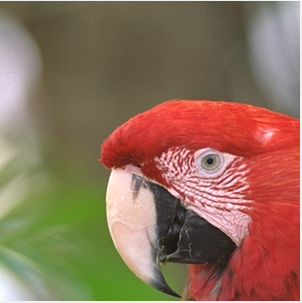}}
\subfigure{
\label{Fig.sub.2}
\includegraphics[width=0.11\textwidth]{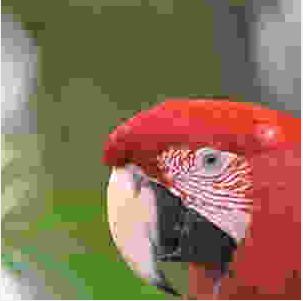}}
\subfigure{
\label{Fig.sub.3}
\includegraphics[width=0.11\textwidth]{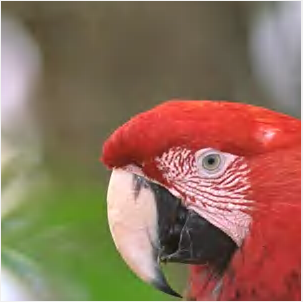}}
\subfigure{
\label{Fig.sub.4}
\includegraphics[width=0.11\textwidth]{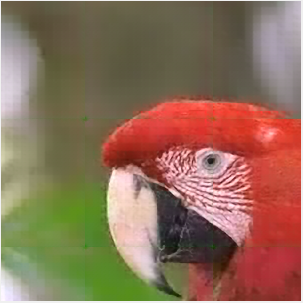}}
\footnotesize{24bpp \qquad\qquad\quad 0.283bpp \qquad\qquad 0.300bpp \qquad \qquad 0.294bpp}
\subfigure{
\label{Fig.sub.1}
\includegraphics[width=0.11\textwidth]{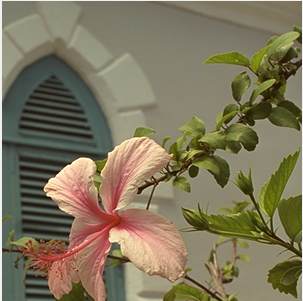}}
\subfigure{
\label{Fig.sub.2}
\includegraphics[width=0.11\textwidth]{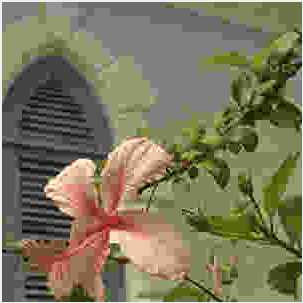}}
\subfigure{
\label{Fig.sub.3}
\includegraphics[width=0.11\textwidth]{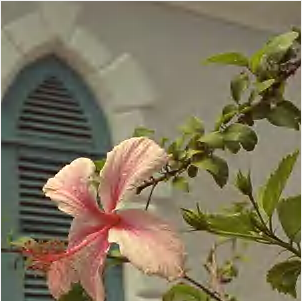}}
\subfigure{
\label{Fig.sub.4}
\includegraphics[width=0.11\textwidth]{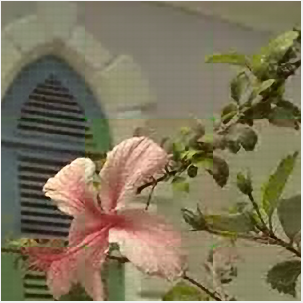}}
\footnotesize{24bpp \qquad\qquad\quad 0.318bpp \qquad\qquad 0.299bpp \qquad \qquad 0.295bpp}\\
\vspace{2pt}
\footnotesize{(a) Raw \qquad\qquad (b) JPEG \qquad\quad(c) JPEG2000 \qquad(d) CAE}
\caption{Examples of raw image (a) and reconstructed images ($300\times300$) cropped from Kodak images using (b)JPEG, (c)JPEG2000 and (d)CAE.}
\label{fig:reconstructed}
\end{figure}

We compare our CAE-based image compression with JPEG and JPEG2000. The color space in this experiment is YUV444. Since the human visual system is more sensitive to the luma component than chroma components, it is common to assign the weights $\frac{6}{8}$, $\frac{1}{8}$, and $\frac{1}{8}$ to the Y, Cb, and Cr components, respectively. The RD curves for the images \emph{red door} and \emph{a girl} are shown in Fig.~\ref{fig:RD}. The coding efficiency of CAE is better than those of both JPEG2000 and JPEG in terms of PSNR. In terms of MS-SSIM, CAE is better than JPEG and comparable with JPEG2000, because optimizing MSE in CAE training leads to better PSNR characteristic, but not MS-SSIM. Besides, CAE handles a fixed input size of $128\times 128$; therefore, block boundary artifacts appear in some images. It is expected that adding perceptual quality matrices into the loss function will improve the MS-SSIM performance, which will be carried out in our future work. Examples of reconstructed patches are shown in Fig.~\ref{fig:reconstructed}. We can observe that the subjective quality of the reconstructed images for CAE is better than JPEG and comparable with that of JPEG2000.

\begin{figure}[tb]
	\centerline{\psfig{figure=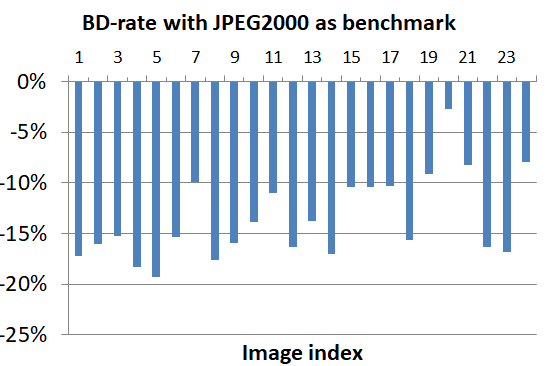,width=56.7mm} }
	\caption{BD-rate of the proposed CAE with JPEG2000 as the benchmark.}
	\label{fig:bdrate}
\end{figure}

The rate-distortion performance can be evaluated quantitatively in terms of the average coding efficiency differences, BD-rate (\%)~\cite{IEEEexample:bdrate}. While calculating the BD-rate, the rate is varied from 0.12bpp to 2.4bpp and the quality is evaluated by using PSNR. With JPEG2000 as the benchmark, the BD-rate results for 24 images in the Kodak database are listed in Fig.~\ref{fig:bdrate}. On average, for the 24 images in the Kodak database, our method achieves 13.7\% BD-rate saving compared to JPEG2000.


\begin{figure}[tb]
\centering
\subfigure[]{
\label{Fig.sub.a1}
\includegraphics[width=0.23\textwidth]{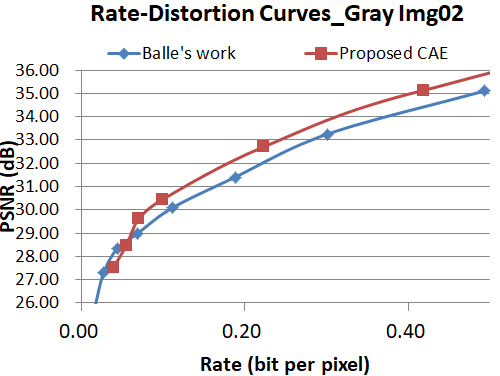}}
\subfigure[]{
\label{Fig.sub.a2}
\includegraphics[width=0.23\textwidth]{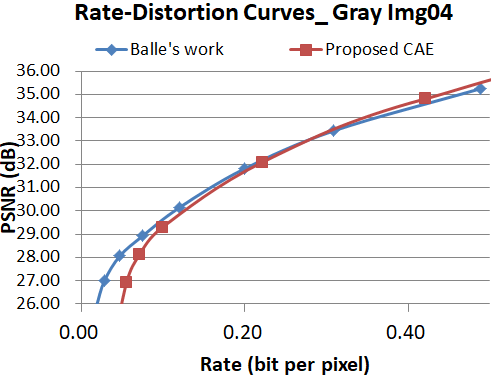}}
\caption{RD curves of gray images for our proposed CAE and Balle's work.}
\label{fig:RD2}
\end{figure}

We also compare our proposed CAE-based method with Balle's work, which released the source code for gray images~\cite{IEEEexample:Balle}. For a fair comparison, we give the comparison results for gray images. For Balle's work, the rate is estimated by the entropy of the discrete probability distribution of the quantized vector, which is the lower bound of the rate. In our work, the rate is calculated by the real file size (kb) divided by the resolution of the tested images. Two examples of RD curves are shown in Fig.~\ref{fig:RD2}. Our method exhibits better RD curves than Balle's work for some test images, such as Fig.~\ref{Fig.sub.a1}, but exhibits slightly worse RD performance for some images, such as Fig.~\ref{Fig.sub.a2}. On average, the performance of our proposed method CAE is comparable with Balle's work, even though the CAE used an actual entropy coder against the ideal entropy of Balle's work.

\subsection{Complexity Performance}

Our experiments are performed on a PC with 4.20 GHz Intel Core i7-7700K CPU, 16GB RAM and GeForce GTX 1080 GPU. The pre-processing steps for the images and Balle's codec~\cite{IEEEexample:Balle} are implemented using Matlab script in Matlab R2016b environment. The codecs of JPEG and JPEG2000 can be found from~\cite{IEEEexample:libjpeg} and~\cite{IEEEexample:Openjpeg}, implemented with CPU. Balle released only their CPU implementation. Running time refers to one complete encoder and decoder process for one color image with a resolution of $768\times 512$, while Balle's time refers to the gray image. The running time comparison for each image for different image compression methods is listed in Table~\ref{Tabel.time}. It can be observed that our CAE-based method achieves lower complexity than Balle's method~\cite{IEEEexample:Balle} when it is run by the CPU, because we have designed a relatively simple CAE architecture. Besides, with GPU implementation, our method could achieve comparable complexity with those of JPEG and JPEG2000, which are implemented by C language. Thus, it proves that our method has relatively low complexity.

\begin{table}[h]
\begin{center}
\caption{Average running time comparison.}
\label{Tabel.time}
\begin{tabular}{|l|l|}
 \hline
 \textbf{Codec} & \textbf{Time (s)} \\
 \hline
 JPEG      & 0.39\\
 \hline
 JPEG2000 & 0.59\\
  \hline
 Balle's work\cite{IEEEexample:Balle} with CPU & 7.39 \\
 \hline
 \textbf{Propose CAE with CPU}  & \textbf{2.29} \\
 \hline
 \textbf{Propose CAE with GPU}  & \textbf{0.67} \\
 \hline
\end{tabular}
\end{center}
\end{table}


\vspace{-2mm}
\section{Conclusion and Future Work}
\vspace{-2mm}

In this paper, we proposed a convolutional autoencoder based image compression architecture. First, a symmetric CAE architecture with multiple downsampling and upsampling units was designed to replace the conventional transforms. Then this CAE was trained by using an approximated rate-distortion function to achieve high coding efficiency. Second, we applied the PCA to the feature maps for a more energy-compact representation, which can benefit the quantization and entropy coder to improve the coding efficiency further. Experimental results demonstrate that our method outperforms conventional traditional image coding algorithms and achieves a 13.7\% BD-rate decrement compared to JPEG2000 on the Kodak database images. In our future work, we will add perceptual quality matrices, such as MS-SSIM or the quality predicted by neural networks in~\cite{IEEEexample:ISM}, into the loss function to improve the MS-SSIM performance. Besides, the generative adversarial network (GAN) shows more promising performance than using autoencoder only; therefore, we will utilize GAN to improve the coding efficiency further.

\end{document}